\documentclass{article}
\usepackage{spconf}
\usepackage{amsmath}
\usepackage{epsfig}
\usepackage{tabularx}
\usepackage{graphicx}
\usepackage{multirow}
\usepackage{bbding}
\usepackage{booktabs}
\usepackage{makecell}
\usepackage{amssymb}
\usepackage{hyperref}
\usepackage{Commands}

\usepackage[dvipsnames]{xcolor}

\newcommand{\numQuestionsAnswers}{N_{Q,A}}

\title{Segmentation-guided Attention for Visual Question Answering from Remote Sensing Images}
\name{\small Lucrezia Tosato\textsuperscript{* a, b} , Hichem Boussaid\textsuperscript{ * a} , Flora Weissgerber\textsuperscript{b} , \textit{Camille Kurtz}\textsuperscript{a} , \textit{Laurent Wendling}\textsuperscript{a}, \textit{Sylvain Lobry}\textsuperscript{a}\thanks{\hspace{-2em} * Lucrezia Tosato and Hichem Boussaid contributed equally.\\This work is supported by \textit{Agence Nationale de la Recherche} (ANR) under the ANR-21-CE23-0011 project.\\The experiments conducted in this study were performed using HPC/AI resources provided by GENCI-IDRIS (Grant 2023-AD011012735R2).}}
\address{\textsuperscript{a}LIPADE, Université Paris Cité, 75006 Paris, France\\ \textsuperscript{b} DTIS, ONERA, Université Paris Saclay, FR-91123 Palaiseau, France}

\begin{document}
%
\maketitle
\begin{abstract}
Visual Question Answering for Remote Sensing (RSVQA) is a task that aims at answering natural language questions about the content of a remote sensing image. The visual features extraction is therefore an essential step in a VQA pipeline. 
By incorporating attention mechanisms into this process, models gain the ability to focus selectively on salient regions of the image, prioritizing the most relevant visual information for a given question.
In this work, we propose to embed an attention mechanism guided by segmentation into a RSVQA pipeline.
We argue that segmentation plays a crucial role in guiding attention by providing a contextual understanding of the visual information, underlying specific objects or areas of interest.
To evaluate this methodology, we provide a new VQA dataset that exploits very high-resolution RGB orthophotos annotated with 16 segmentation classes and question/answer pairs. 
Our study shows promising results of our new methodology, gaining almost 10\% of overall accuracy compared to a classical method on the proposed dataset.
\end{abstract}
\section{Introduction}
\label{sec:introduction}
Visual Question Answering (VQA) is designed to deliver natural language answers to open-ended queries about the content of an image~\cite{antol2015vqa}. Remote Sensing Visual Question Answering (RSVQA) is the application of VQA to remote sensing images, introduced in ~\cite{lobry2020rsvqa}.
Recent advancements in computer vision and natural language processing have elevated the performance of standards on both VQA and RSVQA benchmarks. Notably, Large Language Models have become capable of executing knowledge-based VQA, utilizing external knowledge and commonsense reasoning~\cite{hu2023promptcap}. Large language models have been used in RSVQA and showed great potential to encode the question and to fuse the question with the classes present in the satellite images~\cite{chappuis2022prompt}. Additionally, transformers, such as VisualBERT, have been effectively used to fuse the image and the text~\cite{siebert2022multi}. 
Attention mechanisms have also been used in RSVQA to enhance the features by considering the alignments between spatial positions and words~\cite{zheng2021mutual}.

In the field of computer vision, ~\cite{zhang2020context} and ~\cite{wei2020multi} demonstrate that incorporating attention mechanisms enhances outcomes compared to relying solely on the image. 

In natural images, semantic segmentation has been used to guide visual question answering towards the object of interest~\cite{wu2018faithful}, as well as to guide attention~\cite{Gan_2017_ICCV}.
To the best of our knowledge, these techniques have not yet been used in the field of remote sensing visual question answering. 

Our contribution in this work is to propose to embed a segmentation-guided attention mechanism into a RSVQA pipeline. 
To compute attention weights we consider a  multi-channel semantic segmentation, which differs from traditional one-channel semantic segmentation maps by allowing overlapping of objects of different classes.
We showcase the interest of this methodological contribution on a new RSVQA dataset centered on the Ile-de-France region of France.
It contains very high resolution airborne images, segmentation annotations as well as the automatically constructed question/answer pairs.

The remainder of this article is organized as follows.
The proposed dataset is first described in section \ref{sec:dataset}.
Section \ref{sec:methods} then introduced our novel RSVQA pipeline, including segmentation-guided attention. 
Finally, the results are presented and discussed in section \ref{sec:resultsdiscussion}.

\section{Dataset}
\label{sec:dataset}
\subsection{Very high resolution orthophotos (BD ORTHO) with segmentation annotation}
BD ORTHO is a database of aerial optical images at a resolution of 20cm obtained from the French National Geographic Institute (IGN). The Very High Resolution (VHR) RGB-patches are derived from the subdivision of the BD ORTHO tiles into patches measuring 1000 × 1000 pixels (equivalent to 200m × 200m).

The IGN also provides a vectorial description of the French territory in the BD TOPO database. 
From the latter, we extract 16 classes for our multi-class segmentation annotations: Buildings (Building, Cemetery, Sports Field, Water Tank, Pylon, Surface Construction), Land Use (Foreshore Zone, Vegetation Zone), Water Area, Transport (Airfield, Transportation Construction, Road, Railway), Regulated Areas (Public Forest, National Park), and Services and Activities (that encompasses Museums, Monuments, Schools, etc).

We select four adjacent departments from the Ile-de-France region: 
Paris, Hauts-de-Seine, Val-de-Marne and Seine-Saint-Denis, resulting in 16'274 patches of size $1000\times 1000$ pixels.

\subsection{Question/answer pairs}
Following a procedure similar to \cite{lobry2019visual}, we propose an automated approach to generate sets of questions and answers linked to individual VHR patches. This method fully leverages the BD TOPO database, encompassing both general geographical features, such as buildings and water areas, as well as more specific entities like museums and lakes.

For a given VHR patch $\pvhr$, we retrieve from the BD TOPO the collection of geo-located objects that are present in the geographical extent of $\pvhr$. The objects are caracterized by one element present in BD TOPO that we call a class, e.g. "road".
We define nine question types, divided in four categories: 
\begin{enumerate}
\item  \textbf{Class questions} -- that consider only one class of objects at a time. The questions are divided into  \textbf{presence} (a),  \textbf{count} (b) and  \textbf{density} (c);

\item  \textbf{Objects questions} -- that consider the \textbf{absolute location} in the image (a) or the  \textbf{area} (b) of a specific object;

\item  \textbf{Two-class questions} -- that \textbf{compare} the number of objects of two different classes;

\item \textbf{Object relation questions} -- that consider the \textbf{relative location} (a) and the \textbf{distance} (b) of two specific objects from two different classes, or the absolute location of the \textbf{nearest} object from one class to a pre-selected object from another class or a pixel position (c).
\end{enumerate}

\begin{figure}[t!]
\centering
\includegraphics[width=1\columnwidth, trim={0,1cm 0cm 0,1cm 0,1cm},clip]{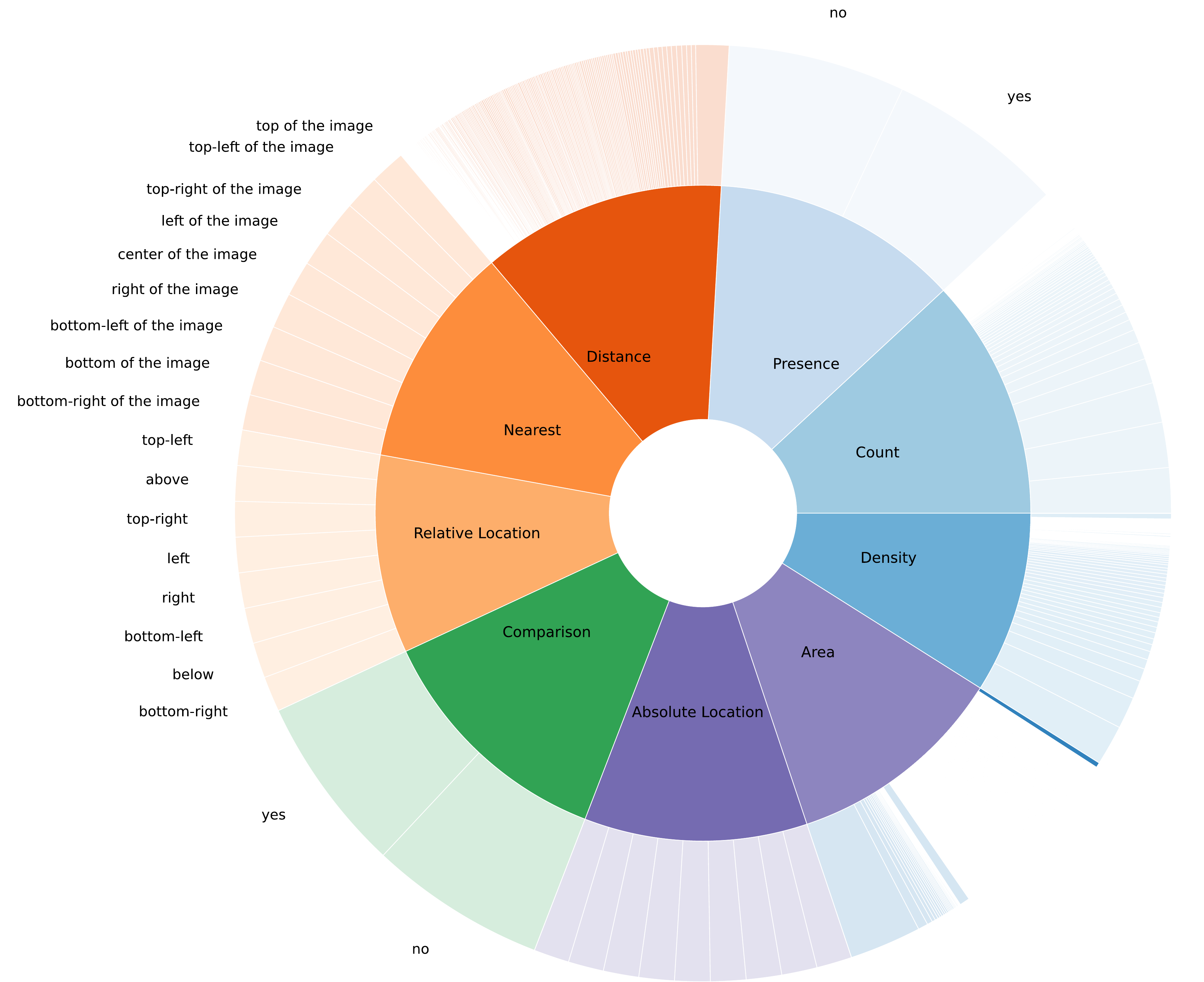}
\caption{Distribution of answers by question type. We omit numerical answers labeling and we show them ordered. The maximum numerical values are 280 (counting questions),  40000m2 (area questions), 273m (distance questions).  }
\label{fig:AnswerDistrib}
\end{figure}

One of the challenges of constructing a VQA dataset stochastically is to balance the question type and the answer type. This is a requirement to reduce language biases~\cite{chappuis2023curse}. 
To balance the question type, we first randomly generate 10 questions per question type for each VHR patch $\pvhr$. 
Then, to balance the answer type, we define a maximum-number-of-questions per answer type $\numQuestionsAnswers$ for each question type. 
Only the questions with an answer type less present than $\numQuestionsAnswers$ are kept. 
To get a sufficient number of questions, the VHR patches are browsed twice. Applying this procedure results in a total of 146'848 question/answer pairs. Their distribution is graphically represented in~\autoref{fig:AnswerDistrib}.

\begin{figure*}[ht]
    \centering
    \includegraphics[width=0.97\textwidth, trim={0 10cm 0 0},clip]{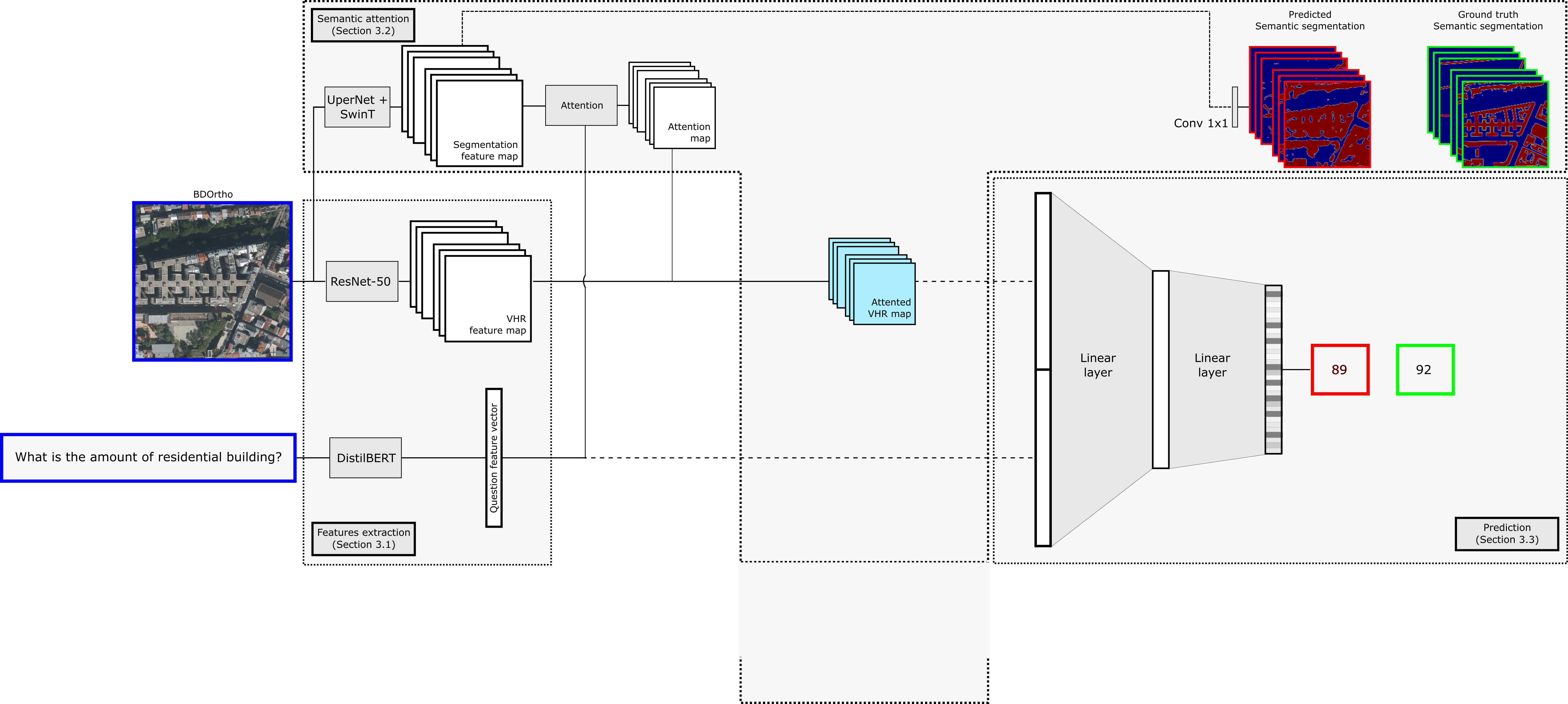}
    \caption{Graphical outline of the proposed architecture. The inputs (very high resolution remote sensing images and language questions) are shown in \textcolor{blue}{blue} frames, the outputs in \textcolor{red}{red} frames (answer and segmentation) and the ground truths (answer and segmentation) in \textcolor{green}{green} frames.}
    \label{fig:method}
\end{figure*}

\subsection{Dataset splitting}
The VHR patches are randomly attributed to the training, validation and test sets with a proportion of 60\%, 20\%, and 20\% respectively.
We use the same dataset splitting for both the segmentation auxiliary task and the complete pipeline.

\section{Method}
\label{sec:methods}

We introduce a novel approach for addressing Remote Sensing Visual Question Answering (RSVQA) by leveraging  segmentation to direct the  attention mechanism. 
The architecture of our pipeline is illustrated in \autoref{fig:method}. 

\subsection{Features extraction}
\label{ssec:feats}

To extract visual features, noted $\fvhr$, from the VHR patches, we use a frozen ResNet-50 model~\cite{he2016deep} pre-trained on ImageNet from which we remove the last fully-connected layer.

The textual features $\fq$ from the question $\q$ are extracted with a DistilBERT encoder~\cite{sanh2019distilbert}, trained on the BookCorpus dataset~\cite{Zhu_2015_ICCV} and frozen in the rest of this study. 

\subsection{Segmentation-guided attention}
\label{ssec:attention}
In our approach, we incorporate an auxiliary per-class segmentation task to guide the computation of the attention weight. 
We assume that the auxiliary task allows a better initialization of the attention module by explicitly introducing a first layer of semantics in the features space.
This allows us to facilitate the learning process of the attention module.

We first train the segmentation module by fine-tuning a UPerNet model with a Swin Transformer backbone introduced in~\cite{liu2021Swin}; pre-trained on ADE20K ~\cite{zhou2019semantic} on the 16 segmentation classes.
To do so, we add a convolutional layer that maps the feature maps $\fseg$ to the 16 channels output.
This output is finally interpolated to the spatial size of $\pvhr$ to obtain the segmentation result.
The weights of the segmentation module are then kept frozen.


We obtain the attention weights by applying a linear layer (with dropout of $d = 0.5$) on $\fq$ to map it to a 250-dimensional space.
Similarly, we use a $1\times 1$ convolution (with a dropout of $d$) to obtain a 250 channels representation of $\fseg$.
We concatenate both representations from the segmentation and the text, and apply a ReLU activation function followed by a 
convolution to obtain one attention glimpse $\att$. 
Finally, we apply $\att$ to the visual features $\fvhr$. 

\subsection{Prediction}
\label{ssec:pred}
We concatenate the attended version of $\fvhr$ with $\fq$ to obtain a global representation of the inputs. This representation is mapped to a 1000-dimensional output (corresponding to the 1000 most frequent answers from the training set) using a 2-layer perceptron. As for the attention, we use the ReLU activation function and a dropout of $d$. We train the model using a cross-entropy loss, optimized with Adam, with a learning rate of $10^{-6}$ and a batch size of 4 samples.

\subsection{Evaluation}

We evaluate the segmentation results with the overall precision / recall and the overall F1-score.

Three metrics are used to evaluate the VQA results: 
the per-question type accuracy, the overall accuracy (OA) and the average accuracy (AA). 
The per-question type accuracy is defined as the ratio of correct answer with the total number of questions for one of the nine question types. 
The OA is the ratio of correct answers with the total number of questions in the dataset. 
Finally, the AA is the average of the per-question type accuracies.

\section{Results and Discussion}
\label{sec:resultsdiscussion}

\label{sec:resultsdiscussioncorr}
\begin{table*}[h]
\centering
\footnotesize
\begin{tabular}{|l|l|l|l|l|l|c|c|c|c|c|c|c|c|c|c|c|} \hline 

\multirow{2}{*}{\textbf{Model}}  &\multicolumn{2}{|c|}{Model param.}&     \multicolumn{9}{|c|}{Per-question accuracy}&\multicolumn{2}{|c|}{Dataset metrics}\\ \cline{2-14}  
   & Att.&Seg.&     1.(a)& 1.(b)& 1.(c)& 2.(a)&2.(b) & 3. &4.(a)& 4.(b)&4.(c)&AA&OA\\ \hline \hline

Proposed&  \checkmark&\checkmark& \textbf{89.36}     &\textbf{26.36}  &\textbf{23.42}&  \textbf{14.61}&\textbf{56.98} &\textbf{93.08} & \textbf{13.76}  & \textbf{74.19} &\textbf{17.21} &\textbf{43.24} &  \textbf{45.44} \\ \hline  

A1    &\checkmark & & 85.79& 22.98 & 20.99 & 13.82 & 44.69 & 82.94 & 11.91 & 74.19 & 15.60 &  39.40 & 41.43 \\ \hline  

A2   & && 80.94  & 10.50& 20.38 & 12.96  & 40.94 & 74.00 & 13.41 &59.80 & 13.44  & 34.91 & 36.26  \\ \hline  

\end{tabular}
\caption{Results of our proposed model and ablation studies. All of the results are accuracy percentages.}
\label{tab:rescor}
\end{table*}


\subsection{Segmentation auxiliary task}
Our model is trained using a NVIDIA GeForce RTX 4090 24G GPU.
Our model achieves an overall precision and recall of 53.6\% and 73.2\% respectively and an overall F1-score of 60.1\%. These results are obtained using a precision-recall curve with 20 different thresholds, to identify on the validation set the optimal threshold value for each class.

\subsection{VQA task}
Our model is trained for 100 hours, using one NVIDIA V100-16G GPU.
The overall results for the VQA task are presented in \autoref{tab:rescor}.
We observe an AA of 43.24\% and an OA of 45.44\% using our segmentation-guided attention.

To study the impact of the segmentation-guided attention we design two ablation studies. 
The first one uses a vanilla attention mechanism (i.e. not guided by the segmentation) and the second one does not use an attention mechanism.

These ablation studies demonstrate that using segmentation-guided attention greatly improves the performances, with a gain of 10\% in overall accuracy and 5\% on the vanilla attention mechanism.

\begin{figure}[t]
    \centering
    \includegraphics[width=1\columnwidth]{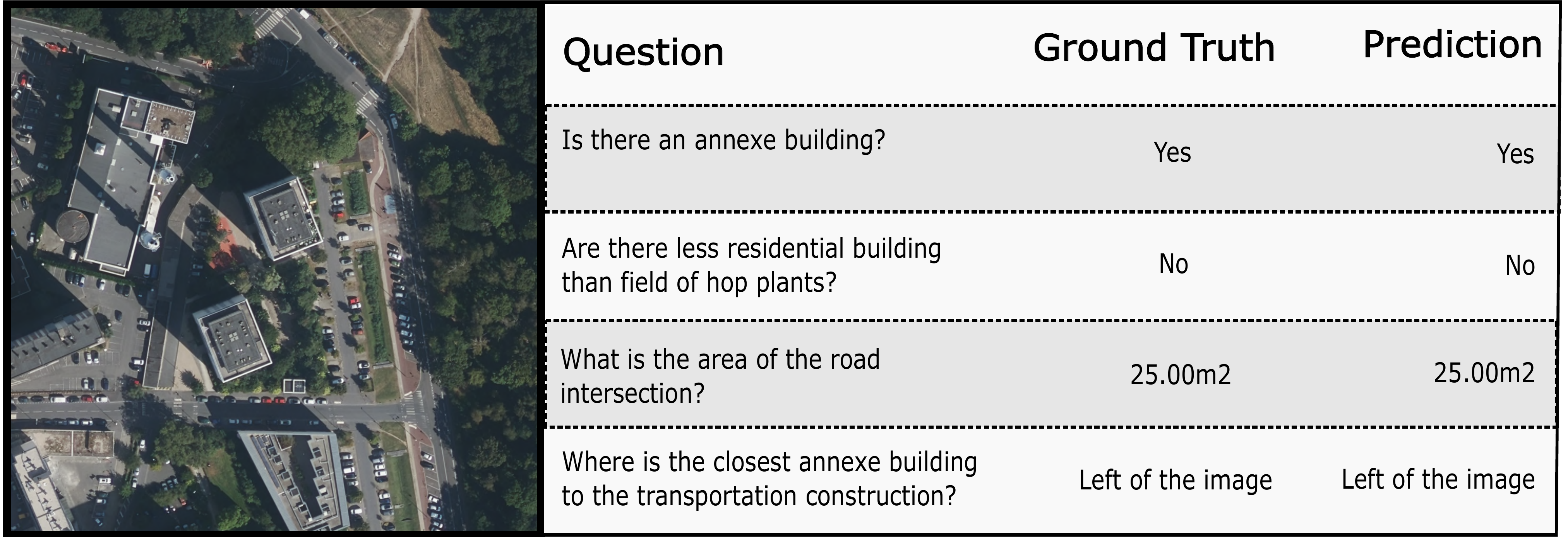}
    \caption{Example of an image in department Hauts-de-Seine, 92 with questions, ground truths and predictions.}
    \label{fig:example}
\end{figure}

A visual example of our model is shown in Figure~\ref{fig:example}.
In table~\ref{tab:rescor} an improvement can be observed in performance using segmentation-guided attention across each of the question types.  In almost all cases, the improvement is gradual, transitioning from A2 to A1 in the proposed method. This demonstrates that the attention mechanism enables the neural network to selectively focus on specific input parts, thereby assisting the RSVQA task. However, the auxiliary task of segmentation increases the level of supervision in the process, leading to better results.

We observe lower prediction accuracy for specific questions (2a, 4a, 4c) that involve determining objects' relative and absolute positions. This challenging task likely contributes to the decreased performance in these cases. It is indeed difficult to simultaneously recognize remote sensing objects and their spatial relationship from end-to-end only relying on present deep learning networks~\cite{cui2019multi}.
To increase the accuracy of such classes, having segmentation maps available, one could use histograms of forces to model the directional spatial relations between geo-localised objects as proposed in~\cite{faure2022embedding}.\\
We can observe that in class density questions (1c), we have a slightly lower accuracy, which is an interesting result when compared to the performances of the area questions (2b). These two tasks, although similar, present a substantial difference: the area prediction is indeed made by numerical values greater than zero and integer, while the density predictions are all values from 0 to 1. 
We believe that the difference in the scale of results imposes a higher sensitivity and precision in the density predictions. 
Another reason might be that the number of density questions is smaller than the others. Finally, while area questions are about one single object, density questions are about a class of objects, complicating the model's task.



\section{Conclusions}
\label{sec:conclusions}
In this study, we apply a segmentation-guided attention model in the context of RSVQA on a new dataset built from both BD ORTHO high-resolution images and BD TOPO for the segmentation annotation, as well as the questions/answers pairs. 

From the preliminary results, we observe that segmentation succeeds in directing attention more effectively than attention alone. 
We believe that by using 16-channel segmentation, attention identifies the channels related to a specific word in the question and thus it is easier to locate the correct object in the final image.
Further experiments with a more complete dataset are necessary to verify that the results can generalize to different geographical areas and more diverse questions. 
\bibliographystyle{ieeetr}
\bibliography{strings,refs}

\end{document}